%% file: main.tex
\documentclass[journal]{IEEEtran}

\usepackage{times}
\usepackage{epsfig}
\usepackage{graphicx}
\usepackage{amsmath}
\usepackage{amssymb}

\usepackage{balance}

\usepackage{bm}
\usepackage{multirow}
\usepackage[dvipsnames]{xcolor}
\newcommand{\matr}[1]{\mathbf{#1}}
\usepackage[pagebackref=true,breaklinks=true,colorlinks,bookmarks=false]{hyperref}

\newcommand{\ie}{\textit{i}.\textit{e}., }
\newcommand{\eg}{\textit{e}.\textit{g}., }

\begin{document}

\title{
Sign Language Recognition via Skeleton-Aware Multi-Model Ensemble
}

\author{Songyao Jiang, Bin Sun, Lichen Wang, Yue Bai, Kunpeng Li and Yun Fu \\
Northeastern University, Boston MA, USA\\

}

%

\maketitle


\begin{abstract}
Sign language is commonly used by deaf or mute people to communicate but requires extensive effort to master. It is usually performed with the fast yet delicate movement of hand gestures, body posture, and even facial expressions. 
Current Sign Language Recognition (SLR) methods usually extract features via deep neural networks and suffer overfitting due to limited and noisy data. 
Recently, skeleton-based action recognition has attracted increasing attention due to its subject-invariant and background-invariant nature, whereas skeleton-based SLR is still under exploration due to the lack of hand annotations. Some researchers have tried to use off-line hand pose trackers to obtain hand keypoints and aid in recognizing sign language via recurrent neural networks. Nevertheless, none of them outperforms RGB-based approaches yet. 
To this end, we propose a novel Skeleton Aware Multi-modal Framework with a Global Ensemble Model (GEM) for isolated SLR (SAM-SLR-v2) to learn and fuse multi-modal feature representations towards a higher recognition rate. Specifically, we propose a Sign Language Graph Convolution Network (SL-GCN) to model the embedded dynamics of skeleton keypoints and a Separable Spatial-Temporal Convolution Network (SSTCN) to exploit skeleton features. 
The skeleton-based predictions are fused with other RGB and depth based modalities by the proposed late-fusion GEM to provide global information and make a faithful SLR prediction. 
Experiments on three isolated SLR datasets demonstrate that our proposed SAM-SLR-v2 framework is exceedingly effective and achieves state-of-the-art performance with significant margins.
Our code will be available at 
\url{https://github.com/jackyjsy/SAM-SLR-v2} 
\end{abstract}

\section{Introduction}
Sign language~\cite{SL_intro} is a visual language performed with the dynamic movement of hand gestures, body posture, and facial expressions. It is a widely-used alternative approach for deaf and mute people to communicate effectively. Understanding and performing sign language requires a substantial time of learning which is beyond feasible for the public, which leads to a barrier between deaf-mute people and others. Moreover, sign language is dependent on language~\cite{SL_book1,SL_book2,yang2010chinese} (\eg English and Chinese) and culture~\cite{different_culture} (\eg American and British) that further limits its popularity. Sign Language Recognition (SLR) aims to help deaf-mute people communicate smoothly with others in their daily life by automatically interpreting sign language. As machine learning and computer vision make great progress in the past decade, SLR has drawn much research attention.

\begin{figure}[t]
\centering
   \includegraphics[width=0.99\linewidth]{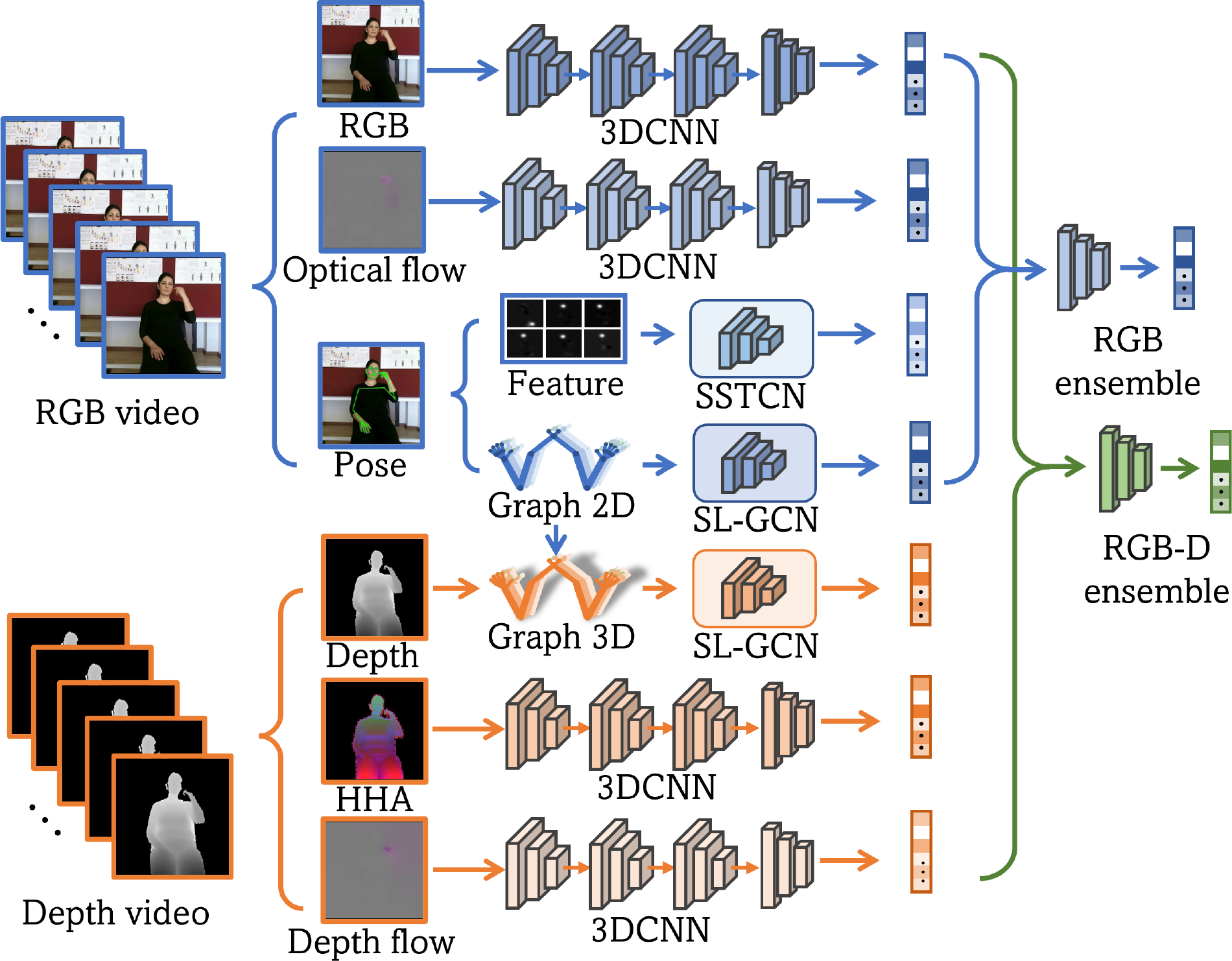}
   \caption{Concept of the Skeleton Aware Multi-modal Sign Language Recognition Framework with Global Ensemble Model (SAM-SLR-v2). All local and global motion information is extracted and fused to make final predictions.}
\label{fig:pipline}
\end{figure}


SLR contains two tasks, isolated SLR and continuous SLR. The isolated setting is a fine-grained and fully supervised classification at word (gloss) level, while the continuous setting maps whole videos into sentences (\ie sequences of glosses) in a weakly supervised manner. 
SLR is a more challenging problem compared with conventional action recognition for the following reasons. 
First, sign language requires both delicate hand gestures and global body motion to distinctly and accurately express its meaning. Facial expressions may also be used to express emotions and emphasis. 
Similar gestures may sometimes express different and even opposite meanings. 
Second, different signers may perform sign language differently (\eg left-hander, right-hander, different speed, body shape, and localism), which makes SLR more challenging. Collecting more samples from as many signers as possible is desired yet expensive. To this end, it requires a finer-grained model to capture the delicate dynamics of the whole human body. Besides, we expect the model to effectively merge the information from different cues when making the final recognition. 

Traditional SLR methods mainly deploy handcrafted features (\eg HOG~\cite{hog_feature} and SIFT~\cite{sift_feature}) with conventional classifiers (\eg kNN and SVM~\cite{yang2010chinese,dardas2011real,memics2013kinect}). 
As deep learning achieves significant progress in video representation learning, general temporal dynamics learning methods (\eg RNN, LSTM, and 3DCNNs) are first explored for SLR in~\cite{sincan2019isolated,li2020word,pigou2018beyond,tur2019isolated}. To more effectively capture the local motion and further improve the accuracy, attention mechanisms are introduced in~\cite{huang2018video,huang2018attention}. Multi-cue approaches are proposed in~\cite{hu2021hand,hosain2021hand,zhou2021spatial} as well. However, due to the extensive vocabulary sizes and the limitation of annotated data, the current methods are still not effective enough for robust SLR.     


Recently, in human action recognition, skeleton-based approaches have become more popular and drawn increasing attention due to their strong adaptability to dynamic circumstances and complicated backgrounds~\cite{yan2018spatial,skeleton_tsrnn,shi2020skeleton,chengdecoupling,song2020stronger}. Besides, the skeleton-based methods provide complementary information to RGB-based modalities, hence their ensemble results further improve the overall recognition rates. 
However, there exist some barriers that hinder the extension of skeleton-based methods to the SLR task. The skeleton-based methods for action recognition rely on ground-truth annotations of human bodies provided by motion capture systems, which restrict themselves to limited datasets captured in lab-controlled environments. Apart from that, most motion capture systems only focus on the fewer body keypoints excluding hand keypoints. The provided keypoints are insufficient to recognize sign language performed with delicate hand gestures and motions. Some researchers attempt to obtain hand poses using separate hand detectors with keypoint estimators and propose to use an RNN-based model to recognize the sign language~\cite{xiao2020skeleton}. Unfortunately, their estimated hand keypoints are unreliable, and the RNN-based model cannot learn the dynamics of the human skeleton properly. 

In this work, we focus on the isolated SLR task. We propose a Skeleton Aware Multi-modal SLR framework with a Global Ensemble Model (SAM-SLR-v2) to explore the potential of skeleton-based SLR and fuse with other modalities in both RGB and RGB-D scenarios to further improve the recognition rate. Specifically, we design a new spatio-temporal skeleton graph for SLR using whole-body keypoints extracted by a pretrained whole-body pose estimator. Then we propose a multi-stream Sign Language Graph Convolution Network (SL-GCN) to model the embedded dynamics. To fully exploit the information in whole-body keypoints, we propose a novel Separable Spatial-Temporal Convolution Network (SSTCN) to learn from the whole-body skeleton features. Moreover, studies on action recognition reveal that data from different modalities complement each other, provide knowledge of latent correlation, and further improve the final performance. Although we can simply add predictions from all modalities together to achieve higher accuracy, we desire a method that tunes the best weight for each modality in a data-driven way. Hence, we propose a Global Ensemble Model (GEM) to automatically learn the multi-modal ensemble and improve the overall recognition rate. 
Our main contributions can be summarized as follows:

\begin{itemize}
\item We construct novel 2D and 3D skeleton graphs designed for SLR using a pretrained whole-body pose estimator and graph reduction, which requires no extra annotation effort. 

\item We propose a novel SL-GCN to model dynamics in the skeleton graphs. To our best knowledge, this is the first successful attempt to tackle the SLR task using 2D/3D whole-body skeleton graphs that surpasses RGB-based methods. 

\item We propose a novel SSTCN to further exploit whole-body skeleton features. It can significantly improve the accuracy compared with the traditional 3D convolution.

\item We propose an ensemble model GEM for both RGB and RGB-D based SLR, which learns weights from seven modalities and achieves state-of-the-art performance on three isolated SLR datasets with significant margins. 

\end{itemize}

A preliminary version of this work~\cite{jiang2021skeleton} has been reported in the corresponding workshop of the SLR challenge~\cite{Sincan:CVPRW:2021}, during which we won the championships in both RGB\footnote{RGB: \url{https://chalearnlap.cvc.uab.es/challenge/43/track/41/result/}} and RGB-D tracks\footnote{RGB-D: \url{https://chalearnlap.cvc.uab.es/challenge/43/track/42/result/}}. Compared with our workshop version, we have made the following improvements: (1) We introduce a new modality Keypoint3D, which considers 3D coordinates in space and deal with occlusions. It improves the overall recognition rate of RGB-D ensemble. (2) We propose a new learning-based late-fusion ensemble method named GEM for multi-modal ensemble, which achieves higher recognition rates than our previous version and saves efforts on weights tuning. (3) The test labels of the AUTSL dataset has been released, so we update our performance from the validation set to the test set. (4) Beyond the challenge dataset (AUTSL), we report our performance on two additional large-scale datasets for isolated (\ie SLR500 and WLASL2000) compared with the recent state-of-the-art methods. Our approach significantly surpasses their performance with notable margins. (5) We update our figures, provide more details of models, analyze ensemble sensitivity, and discuss challenging cases, which may inspire future research on SLR. 

\begin{figure*}[!t]
  \centering
  \includegraphics[width=0.98\textwidth]{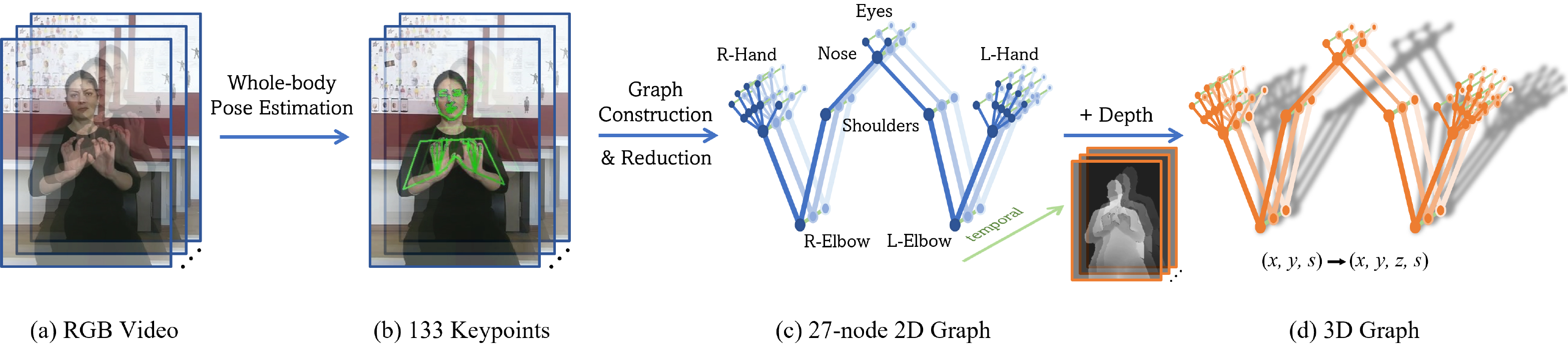}
  \caption{Construction of 2D and 3D graphs. A pretrained whole-body pose estimator is applied on an input RGB video (a) to obtain 133-point whole-body keypoints (b). We then construct a spatio-temporal 2D graph following the natural connections of the human body. The graph is reduced to 27-node (c) using graph reduction to mitigate noises. Each node in the 2D graph is represented by $(x,y,s)$ where $x$-$y$ are 2D coordinates and $s$ is the confidence score. The 3D graph is obtained by overlaying the 2D coordinates on the depth video, reading the depth encoding $z$ at the corresponding location, and treating it as an additional dimension as $(x,y,z,s)$. }
\label{fig:graph-construction}
\end{figure*}

\section{Related work}
\noindent\textbf{Sign Language Recognition (SLR)} achieves significant progress in obtaining high recognition accuracy in recent years due to the development of practical deep learning architectures and the surge of computational power~\cite{huang2018attention,sincan2019isolated,tur2019isolated,lim2019isolated}. 
Researchers have been modeling the spatio-temporal information in the videos using 2D CNN with RNN or 3D CNN. 
A 3D-convolutional SLR network associated with attention modules is proposed in~\cite{huang2018attention} to learn the spatio-temporal features from raw videos. 
CNN, Feature Pooling Module, and LSTM Networks associated with adaptive weights are utilized in~\cite{sincan2019isolated} to obtain distinctive representations.
Besides, researchers have been extending isolated SLR to weakly supervised continuous SLR and Sign Language Translation (SLT) by incorporating sequence learning methods~\cite{koller2018deep,guo2018hierarchical,huang2018video,pu2019iterative,cui2019deep}. 
Recently, multi-cue methods using upper-body poses, hand keypoints, and mouth features have been developed to improve the recognition rate. For example, hand pose priors are introduced in hand-aware frameworks for further performance improvement in~\cite{hosain2021hand,hu2021hand}. A spatial-temporal multi-cue network for continuous SLR and SLT is proposed in~\cite{zhou2021spatial}.
However, these methods are still not effective enough to capture the complete motion information for robust sign language recognition. 

\noindent\textbf{Skeleton-based Action Recognition} mainly focuses on learning dynamic patterns from the motion of the human skeleton and effectively recognizing human action and activities~\cite{cai2021jolo, du2015hierarchical, huang2017deep, li2019actional, li2018independently, liu2016spatio}. In the meanwhile, aiming for a higher recognition accuracy, it can collaborate with other modalities (\eg RGB, depth, and EMG) and benefit from multi-modal learning~\cite{baradel2017human, carreira2017quo, choutas2018potion, hu2018deep, zolfaghari2017chained}. 
Recurrent neural networks (\eg RNN and LSTM) are once popular in modeling the temporal information of skeleton data~\cite{du2015hierarchical, liu2016spatio, li2018independently, si2018skeleton}.
Recently, a graph-based approach that implements a Graph Convolutional Network (GCN) to model the dynamic patterns in skeleton data is first explored by ST-GCN~\cite{yan2018spatial}. It draws much research attention that a few improved models are developed to further improve the performance~\cite{li2019actional, shi2019skeleton, shi2019two, si2019attention, shi2020skeleton, chengdecoupling, song2020stronger}. Specifically,
AS-GCN~\cite{li2019actional} digs the latent joint connections to boost the recognition performance. A two-stream approach is presented in \cite{shi2019two} and further extended to four streams in~\cite{shi2020skeleton}. DecoupleGCN~\cite{chengdecoupling} increases the GCN capacity while introducing no extra computational cost. Inspired by ResNet~\cite{he2016deep}, ResGCN~\cite{song2020stronger} incorporates a bottleneck architecture to boost model capacity, reduce parameters, and avoid gradient vanishing. 
However, skeleton-based SLR methods are still under exploration. Simply applying the ST-GCN to SLR has been unsuccessful, which only reaches around 60\% top-1 accuracy on 20 classes (much worse than RGB-based approaches)~\cite{de2019spatial}. 

\label{sec:related_work}
\noindent\textbf{Multi-modal Approach} aims to explore data captured from different resources, by different devices, and from distinctive views to improve the overall performance. Its motivation lies in an assumption that different modalities contain unique and view-specific information that complements each other. The assumption suggests that multi-modal information can be fused together to further boost performance. 
Frameworks that learn robust and view-invariant feature representation for downstream tasks are proposed in~\cite{zheng2016cross,wang2013dense}. A weight-sharing model is developed in~\cite{hoff_CVPR16} to obtain modality hallucination for multi-modal image classification. DA-Net~\cite{Wang_2018_ECCV} adopts a view-independent module collaborated with a view-specific module to capture the multi-modal information and effectively merges the prediction scores. 
PM-GANs~\cite{pm_gan} utilizes a cross-view generative strategy associated with a novel fusion to learn the prediction correlations effectively. 
A feature factorization network is proposed in~\cite{shahroudy2018deep} which learns the specific view-shared information for RGB-D action recognition. A cascaded residual autoencoder is designed to handle incomplete view classification settings~\cite{tran_CVPR17}. Researchers also propose to use a super vector to fuse the multi-modal representations~\cite{multiview_action2}. 
Encouraged by the success of those multi-modal approaches, we aim to incorporate more modalities (\eg visual, depth, body gesture, and hand gesture) to capture both view-dependent and view-specific information from all aspects. We aim to learn via a universal framework to achieve higher performance.

\begin{figure*}[!ht]
  \centering
  \includegraphics[width=0.92\textwidth]{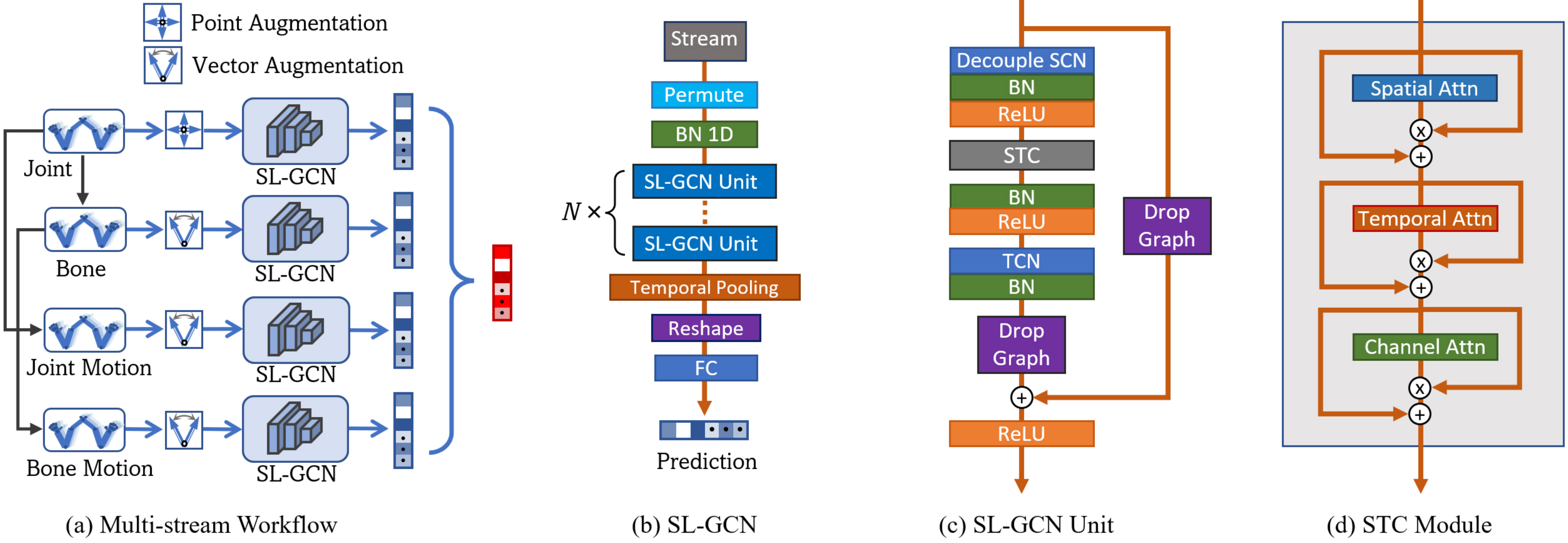}
  \caption{Illustration of the SL-GCN pipeline: (a) The multi-stream workflow includes streams of joint, bone, joint motion, and bone motion; (b) Illustration of the SL-GCN architecture; (c) Network details of the SL-GCN Unit; (d) The STC self-attention module used in the SL-GCN unit consists of a spatial attention module, a temporal attention module, and a channel attention module that connect in a cascaded way.}
\label{fig:gcn-pipeline}
\end{figure*}

\section{Methodology}
This section first gives an overview of the proposed SAM-SLR-v2 framework. Then we introduce the SL-GCN for keypoint graphs and the SSTCN for skeleton features, respectively. After that, we present an effective 3DCNN baseline model for the other modalities. Last, we describe the late-fusion GEM, for the multi-modal ensemble.

\subsection{SAM-SLR-v2 Framework Overview}
An overview of the proposed SAM-SLR-v2 is shown in Figure~\ref{fig:pipline}. Seven modalities processed from RGB and depth videos are considered in the proposed framework. Three different architectures (\ie SL-GCN, SSTCN, and 3DCNN) are used to extract features from the seven modalities and make predictions of sign language glosses independently. The late-fusion ensemble model (\ie GEM) takes predictions from all the modalities and outputs final predictions in the RGB and the RGB-D scenarios. 

\subsection{SL-GCN for Skeleton Keypoints}

\noindent\textbf{Graph Construction and Reduction}. 
\label{sec:graph_construction}
For skeleton-based action recognition, researchers rely on ground-truth skeleton keypoints annotated by motion capture systems such as Kinect v2~\cite{kinect_v1v2}, which unfortunately is not capable of providing hand and finger annotations. Since hand gestures play a crucial role in performing sign language, we use a pose estimator pretrained with whole-body annotations to predict whole-body keypoints, which include 133 landmarks of face, body, hands, and feet. A spatial 2D graph can then be constructed by connecting every pair of adjacent keypoints according to the human-body natural connections. This graph is further extended to a spatio-temporal graph by connecting all the nodes to themselves in the temporal dimension. Mathematically, the node set $V=\{v_{i,t}|i=1,...,N,t=1,...,T\}$ consists of all 133 whole-body nodes. Their adjacent matrix $\matr{A}$ is defined as
\begin{equation}
    \matr{A}_{i,j}=
        \begin{cases}
      1 & \text{if $d(v_i, v_j)=1$}\\
      0 & \text{else}
    \end{cases} 
    \label{eqn:graph_construct}
\end{equation}
where $d(v_i, v_j)$ calculate the minimum distance (\ie the minimum number of nodes) between Node $v_i$ and $v_j$ in the shortest path. 

However, contrasting with the graph used in skeleton-based action recognition which contains around 17 nodes, the whole-body skeleton graph contains too many nodes and edges which introduce high-level unexpected noise. Besides, if the distance between two nodes is too far (\ie have many nodes in between), it is inaccurate to explore their interactions. Our experiment shows that simply using the whole-body skeleton graph results in lower accuracy. Therefore, we operate a graph reduction on the whole-body skeleton graph which trims down the 133 nodes to 27 nodes based on our observations of the videos and visualizations of the GCN activation. The resulted graph consists of seven nodes for the upper body (nose, eyes, shoulders, and elbows) and ten nodes for each hand, as illustrated in Figure~\ref{fig:graph-construction}(c). Graph reduction leads to faster model convergence and significantly higher recognition rates. Each node in the 2D graph is represented by $(x,y,s)$ where $x$-$y$ are 2D coordinates and $s$ is the confidence score. When depth information is available, we construct a 3D graph by reading the corresponding depth $z$ at keypoint locations $x$-$y$ and treating it as an additional dimension as $(x,y,z,s)$, as illustrated in Figure~\ref{fig:graph-construction}(d).  



\noindent\textbf{Graph Convolution}. 
We adopt the spatio-temporal graph convolution with the spatial partitioning strategy~\cite{yan2018spatial} to capture the embedded dynamics in the whole-body skeleton graph. We implement the spatial graph convolution as
\begin{equation}
    \matr{x}_\text{out}=\matr{D}^{-\frac{1}{2}}(\matr{A}+\matr{I})\matr{D}^{-\frac{1}{2}}\matr{x}_\text{in}\matr{W},
    \label{eqn:graph_conv}
\end{equation}
where $\matr{A}$ represents an adjacent matrix of intra-body connections, $\matr{I}$ represents an identity matrix of self-connections, $\matr{D}$ stands for the diagonal degree of $(\matr{A}+\matr{I})$, and $\matr{W}$ is the trainable weights of the convolution. Practically, Equation~\ref{eqn:graph_conv} is implemented by performing standard 2D convolution and multiplying the results by $\matr{D}^{-\frac{1}{2}}(\matr{A}+\matr{I})\matr{D}^{-\frac{1}{2}}$. To perform temporal graph convolution, we implement a standard 2D convolution on the temporal dimension with a kernel size $k_t \times 1$ as the reception field. 
We adopt an extended variation of the spatial GCN called decoupling GCN~\cite{chengdecoupling}. In a decoupling GCN layer, to further boost the model capacity, the extracted features are grouped into $G$ groups that each group has its trainable adjacent matrix $\matr{A}$. We then concatenate the outputs of all groups back together as the output features. 

\noindent\textbf{Multi-stream SL-GCN}. 
\label{sec:multi-stream}
Inspired by \cite{shi2020skeleton} which adopts a multi-stream workflow for action recognition, we find that it is also worth investigating 1st-order coordinates (joints), 2nd-order vector (bone vector), and their motion vectors for sign language recognition, as shown in Figure~\ref{fig:gcn-pipeline}(a). 
Following the natural connections of the human body, we generate the bone vectors pointing from the starting joints to their ending joints. Thus the bone stream is represented by a tree graph where the nose acts as the root node. Mathematically, the starting and ending joints can be represented as 
\begin{equation}
v^\text{J}_{i,t}=(x_{i,t},y_{i,t},[z_{i,t}],s_{i,t}),
\label{eqn:start_joint}
\end{equation}
\begin{equation}
v^\text{J}_{j,t}=(x_{j,t},y_{j,t},[z_{j,t}],s_{j,t}),
\label{eqn:end_joint}
\end{equation}
where $(x,y,[z],s)$ represents 2D coordinates, optional depth, and the confidence score. The bone vectors $v^\text{B}$ are calculated by subtracting Equation~\ref{eqn:start_joint} from Equation~\ref{eqn:end_joint} as
\begin{equation}
v^\text{B}_{j,t}=(x_{j,t}-x_{i,t},y_{j,t}-y_{i,t},z_{j,t}-z_{i,t},s_{j,t}),\forall(i,j)\in \mathbb{H}
\end{equation}
where the set $\mathbb{H}$ contains all natural connections of the human body. Motion streams are obtained by subtracting the difference between adjacent frames. The joint motion vectors $v^\text{JM}$ and the bone motion vectors $v^\text{JM}$ are represented as 
\begin{equation}
v^\text{JM}_{i,t}=v^\text{J}_{i,t+1}-v^\text{J}_{i,t},
\end{equation}
\begin{equation}
v^\text{BM}_{i,t}=v^\text{B}_{i,t+1}-v^\text{B}_{i,t}.
\end{equation}
We train every stream separately, multiply their predictions by assigned weights, and summing up the results as the final prediction.

\noindent\textbf{SL-GCN Structure}. 
The structure of our proposed SL-GCN is presented in Figure~\ref{fig:gcn-pipeline}(b). The input stream is permuted and normalized before feeding into $N$ instances of SL-GCN units for spatio-temporal graph convolution. An average pooling is then applied to the temporal dimension of the resulted features. The result is reshaped and fed into a fully connected layer for classification. 
Our proposed SL-GCN unit is illustrated in Figure~\ref{fig:gcn-pipeline}(c). 
We find that deep graph models are easier to overfit on the video classification task and the ordinary dropout layer works poorly in GCNs. We propose to construct the basic SL-GCN Unit with a decoupled spatial convolutional layer (DecoupleSCN)~\cite{chengdecoupling} to mitigate overfitting. We also introduce an STC (spatial, temporal, and channel-wise) self-attention mechanism inspired by~\cite{shi2020skeleton}. As illustrated in Figure~\ref{fig:gcn-pipeline}(d), the STC module consists of modules of spatial attention, temporal attention, and channel attention connected in a cascaded configuration. In our experiment, we use $N=10$ such SL-GCN units in the proposed SL-GCN.

\begin{figure}[t]
  \centering
  \includegraphics[width=0.42\textwidth]{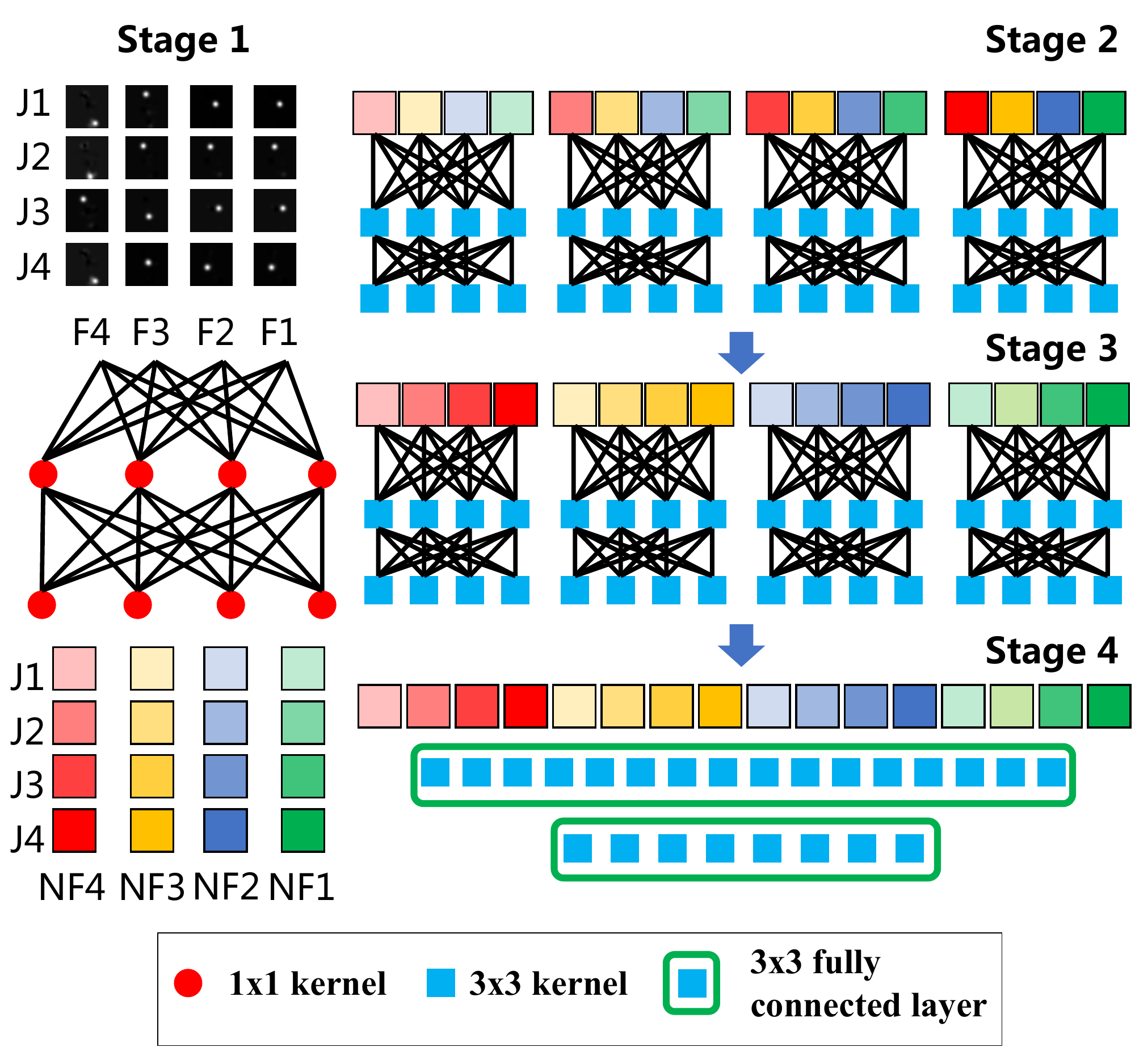}
  \caption{The architecture of SSTCN for skeleton features. Abbrevs: J=Joints; F=Frames; NF=New Features.}
  \label{fig:tpose}
\end{figure}

\subsection{SSTCN for Skeleton Features}
\label{sec:SSTCN}
We propose an SSTCN model to exploit the whole-body features in addition to the keypoint coordinates. We can learn from ResNet2+1D~\cite{tran2018closer} that the performance of an action recognition model can be further improved via factorizing the network into a temporal part and a spatial part. Therefore, our SSTCN model is separated into four stages to handle the features from different dimensions. The pipeline is shown in Figure~\ref{fig:tpose}. We save the features of 33 keypoints including 1 nose keypoint, 4 mouth keypoints, 6 upper-body keypoints, and 22 hands keypoints before the argmax operation in the pose estimator. We then uniformly sample 60 frames in each video for SSTCN for recognition. The saved features are resized to $24\times24$ using maximum pooling. We process the input features with a 2D separable convolution layer that reduces the parameters and converges easily. At Stage one, features are reshaped from $60\times33\times24\times24$ to $60\times792\times24$, and fed to $1 \times 1$ convolution layers for temporal convolution. 
At Stage two, the extracted features are then shuffled and grouped into $60$ groups for $3 \times 3$ grouped convolution to extract spatial features of each frame. At the next stage, we start processing the features in the feature dimension. Specifically, the features are shuffled again and grouped into $33$ groups. We use grouped $3 \times 3$ convolution on spatial and temporal dimensions for each keypoint. At the last stage, a few fully connected layers are used to make final predictions. We adopt a residual path for the first three stages to avoid gradient vanishing. Besides, random dropouts are deployed in every module to avoid overfitting~\cite{dropout}. 

Researchers have found that one-hot labels and cross-entropy loss may be easy to overfit with limited data~\cite{ren2015faster}. So we use the technique of label smoothing to avoid overfitting. Mathematically, label smoothing is defined as 
\begin{equation}
    q'(k|x)=(1-\epsilon)\delta_{k,y}+\epsilon u(k),
    \label{eqn:label_smooth}
\end{equation}
where $k$ is the number of classes, $q'(k|x)$ is a label-smoothed predicted distribution, $\epsilon$ is a hyper-parameter in $(0,1)$, and $u(\cdot)$ represents the uniform distribution. The cross-entropy loss is then modified as
\begin{equation}
    H(q',p)=-\sum_{k=1}^K \log p(k)q'(k) = (1-\epsilon) H(q,p) + \epsilon H(u,p),
    \label{eqn:cross_entropy}
\end{equation}
where $H(\cdot)$ is the cross-entropy function and $q$ is the real data distribution, and $p$ stands for the predicted distribution. The modified cross-entropy can be explained as the penalty to the difference between the predicted distribution and the real distribution combined with the difference between the predicted distribution and a prior distribution (\eg uniform distribution). To further improve the performance, we replace all activations to the Swish~\cite{swish} activation function as
\begin{equation}
    f(x)=x \cdot \text{Sigmoid}(x).
    \label{eqn:swish}
\end{equation}

\begin{figure}
  \centering
  \includegraphics[width=0.45\textwidth]{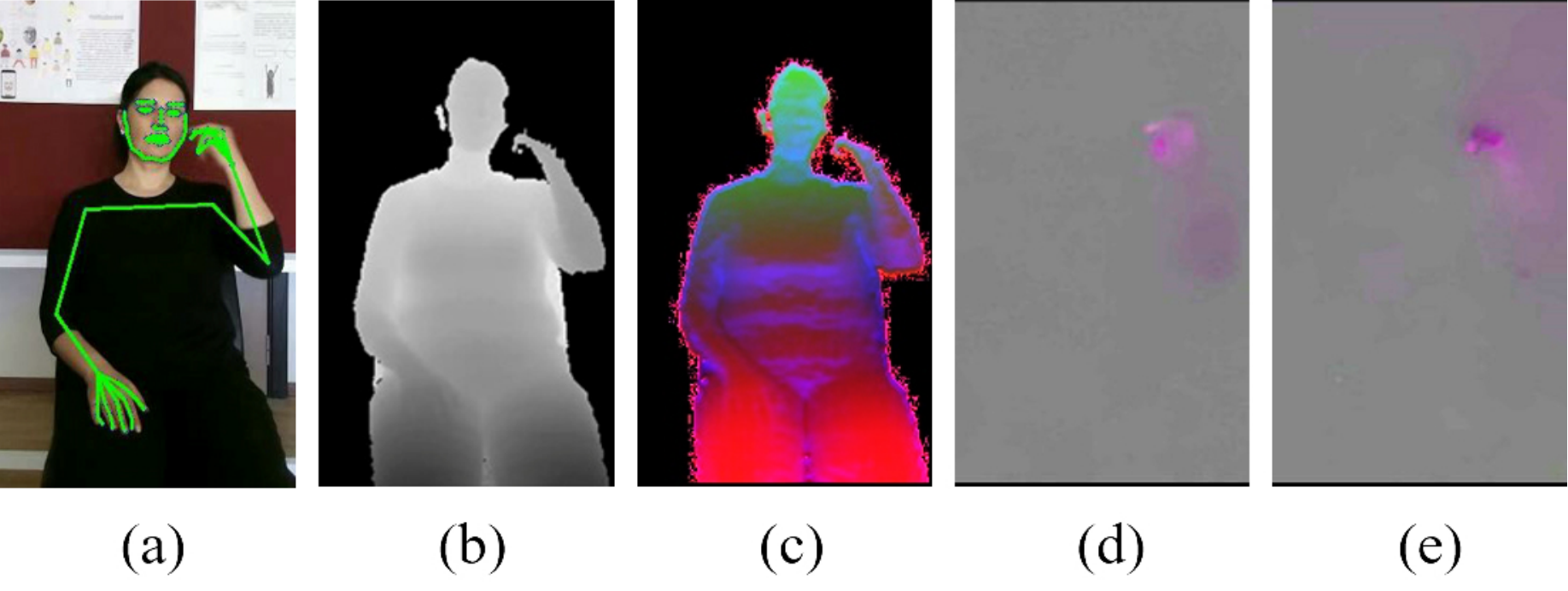}
  \caption{Visualization of modalities: (a) RGB with whole-body keypoints overlay; (b) Depth; (c) Masked HHA; (d) Optical flow; (e) Depth flow. (better viewed in color)}
\label{fig:modalities}
\end{figure}

\subsection{3DCNN Baselines for the Other Modalities}
As mentioned in Section~\ref{sec:related_work}, studies reveal that ensembles from multiple cues and modalities could further improve the overall performance. To benefit from the other modalities (\ie RGB frames, optical flows, HHA, and depth flow), we hence build a simple yet effective baseline using 3D CNNs.  
Most 3D CNN architectures are easy to overfit, especially on smaller datasets. In our experiment, The ResNet2+1D~\cite{tran2018closer} architecture that separates the temporal and spatial convolution of 3D CNNs, provides the best result compared with other popular 3D CNN networks. We also notice that deeper model depth does not always lead to better performance while making the network to overfit easier. So we choose the ResNet2+1D-18 variation pretrained on the Kinectics dataset~\cite{carreira2018short} as our backbone. Apart from that, to improve the accuracy further, we pretrain the model on the largest available SLR dataset SLR500~\cite{zhang2016chinese} for RGB frames. Pretraining increases the final accuracy by around $1\%$ and improves the model convergence. Similar to the SSTCN presented in Section~\ref{sec:SSTCN}, we adopt the Swish activation described in Equation~\ref{eqn:swish} instead of using ReLU. Besides, to mitigate overfitting, we implement the label smoothing technique in Equation~\ref{eqn:label_smooth} with the corresponding cross-entropy loss in Equation~\ref{eqn:cross_entropy}. 

\begin{figure}[t]
  \centering
  \includegraphics[width=0.5\textwidth]{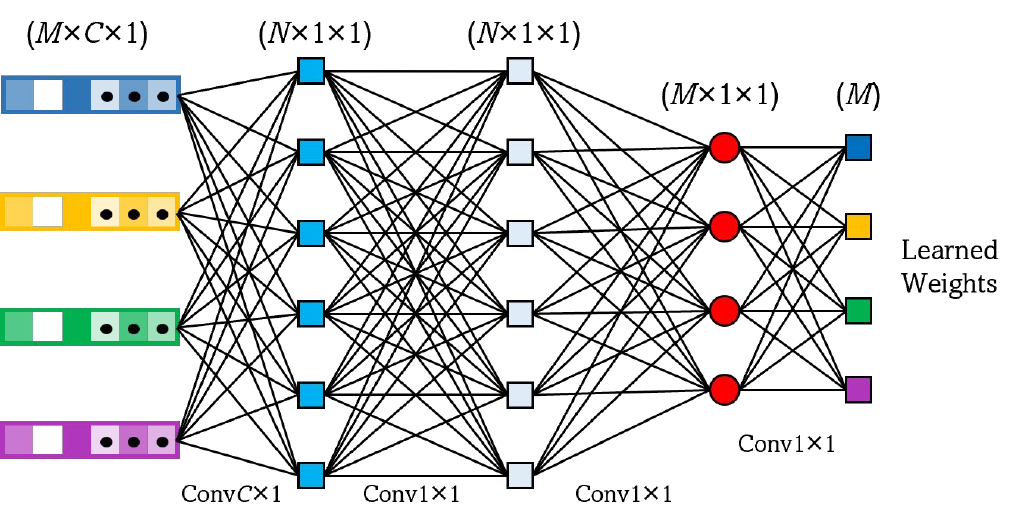}
  \caption{Illustration of the global ensemble model. The predictions of $M$ modalities are concatenated together, then fed into a convolutional layer with kernel size $C\times1$, where $C$ is the number of classes. After a few $1\times1$ fully connected layers with filter size $N$ and layer normalization, we obtain $M$ weights to fuse the modalities together as the final prediction. }
  \label{fig:ensemble}
\end{figure}

\subsection{Multi-modal Late-fusion Ensemble}
\label{sec:ensemble}
\noindent\textbf{Model-free Late Fusion}. In our previous version, we use a simple late-fusion approach to fuse predictions from all modalities. To be specific, for all modalities, we save the predictions from the last fully connected layers before the softmax. We assign weights to all modalities manually and add them up as the final prediction
\begin{equation}
    q_\text{RGB} = \alpha_1 q_\text{skel2D} + \alpha_2 q_\text{RGB}  + \alpha_3 q_\text{flow} + \alpha_4 q_\text{feat},
\end{equation}
\begin{align}
\begin{aligned}
    q_\text{RGB-D} = & \alpha_1 q_\text{skel3D} +  \alpha_2 q_\text{RGB}  + \alpha_3 q_\text{flow} + \alpha_4 q_\text{feat}\\
    &  + \alpha_5 q_\text{HHA} + \alpha_6 q_\text{depthflow},
\end{aligned}
\end{align}
where $q$ represents the predictions before softmax, $\alpha_{1,2,3,4,5,6}$ are hyper-parameters to be tuned based on validation accuracy. In our previous version, we use $\bm{\alpha} = \{1,0.9,0.4,0.4\}$ for RGB track and $\bm{\alpha} = \{1.0,0.9,0.4,0.4,0.4,0.1\}$ for RGB-D track. For the other datasets without validation and test splits, we keep those weights unchanged without further hard-tuning.

\noindent\textbf{Global Ensemble Model}.  
Since finding the best weights for fusion is time-consuming, we propose a learning-based global ensemble model (GEM) to fuse all the modality automatically, as illustrated in Figure~\ref{fig:ensemble}. The whole process can be described as 
\begin{align}
\begin{aligned}
    \{\alpha_i\}_{i=1}^4 = & G_{\text{RGB}}([q_\text{Key2D},q_\text{RGB},q_\text{flow},q_\text{feat}]),\\
    \{\alpha_i\}_{i=1}^6 = &  G_{\text{RGB-D}}([q_\text{Key3D},q_\text{RGB},q_\text{flow},q_\text{feat},q_\text{HHA},q_\text{depthflow}]),
\end{aligned}
\end{align}
where $q$ represents the results of each modality, $G$ represents the prediction procedure of GEM, and $[ \cdot ]$ stands for the concatenation operation. The first layer of the model is used to down-sample the modalities effectively with a global convolution. Therefore, the kernel of the first layer is $C\times1$. Then we use $1 \times 1$ fully connected layers to predict the final weights of all modalities. The final weights are multiplied with the inputs respectively and the weighted modalities are summed up as the final prediction.

\input{experiment}
\section{Conclusion}


In conclusion, we propose a novel SAM-SLR-v2 framework to learn multi-modal feature representations from RGB-D videos towards more effective and robust isolated SLR. 
Among those modalities, our proposed skeleton-based methods are the most effective in modeling motion dynamics due to their signer-independent and background-independent characteristics. 
Specifically, we construct novel 2D and 3D spatio-temporal skeleton graphs using pretrained whole-body keypoint estimators and propose a multi-stream SL-GCN to model the embedded motion dynamics. Our method does not require any additional effort on annotating hands and provides more reliable hand keypoint estimation than off-line hand detectors. Besides modeling motion dynamics of keypoint coordinates via graphs, we propose SSTCN to predict using skeleton features. Furthermore, we study the multi-modal fusion problem based on the other modalities (\ie RGB frames, optical flow, HHA, and depth flow) via a learning-based late-fusion ensemble model named GEM. Experimentally, we show that our proposed SAM-SLR-v2 framework achieves the state-of-the-art performance on three challenging datasets for isolated SLR (\ie AUTSL, SLR500, and WLASL2000) as well as won the championships in both RGB and RGB-D tracks during the CVPR 2021 challenge on isolated SLR. We hope our work could facilitate and inspire future research on SLR.

    

\bibliographystyle{IEEEtran}
\bibliography{ref}


\end{document}

%% file: experiment.tex
\section{Experiments}
In this section, we report the evaluation results of our proposed SAM-SLR-v2 on three major sign language recognition datasets of different sign languages. We show the single-modality performance of our proposed multi-stream SL-GCN, SSTCN, and 3DCNN models, as well as different combinations of their ensembles. We demonstrate the effectiveness of proposed approaches via ablation studies on the components of SL-GCN, SSTCN and 3DCNN networks. We also study the proposed GEM compared with simple model-free late fusion. Last, we show some challenging cases and discuss the limitations of SAM-SLR-v2.

\begin{table}[!t]
\caption{A statistical summary of SLR datasets.\\ * No longer publicly available\\ $\dagger$ Not released at the moment.}
\label{dataset_table}
\centering
\begin{tabular}{l | c c r c}
\hline
Datasets & \#Signs & \#Signers & \#Samples & Languages \\
\hline
AUTSL~\cite{SLR_dataset}            & 226   &  43 & 21,083  & Turkish\\
SLR500~\cite{zhang2016chinese}      & 500   &  50 & 125,000 & Chinese\\
WLASL2000~\cite{li2020word}         & 2,000 & 119 & 21,083  & American\\
MS-ASL~\cite{vaezijoze2019ms-asl} * & 1,000 & 222 & 25,513  & American\\
BSL-1K~\cite{albanie2020bsl} $\dagger$& 1,064 & 40  & 273,000 & British\\
\hline
\end{tabular}
\end{table}

\subsection{Datasets}
\noindent\textbf{AUTSL Dataset}~\cite{SLR_dataset} is a Turkish SLR dataset collected using Kinect V2 sensor~\cite{kinect_v1v2,kinect2_acc}. Statistically, 226 different sign glosses are performed by 43 signers with 20 backgrounds. The dataset contains 38,336 videos that split into training, validation, and testing subsets. We use their currently released balanced test set in our experiments and report both top-1 and top-5 recognition rates.

\noindent\textbf{SLR500 Dataset}~\cite{zhang2016chinese} is a balanced Chinese sign language dataset for isolated sign language recognition (sometimes referred to as CSL isol.) that contains 500 words performed by 50 signers. Each word is performed by all 50 signers 5 times, so there are 125,000 videos in total. This dataset is captured in a controlled lab environment with a solid-color background. The former 36 signers are used for training and the later 14 signers are used for testing.

\noindent\textbf{WLASL2000 Dataset}~\cite{li2020word} is a American Sign Language with a vocabulary size of 2000 words. It is a challenging dataset collected from web videos performed by 119 signers. It contains 21,083 samples with unconstrained recording conditions. The dataset is imbalanced and the average samples per video are much lower than the above two datasets.

We follow the signer-independent settings for all the above datasets. Besides, we are also aware of other large-scale isolated sign language datasets. Microsoft American Sign Language dataset (MS-ASL)~\cite{vaezijoze2019ms-asl} is an American sign language dataset that is no longer publicly available due to the deletion of the YouTube-hosted videos. A new large-scale British Sign Language dataset (BSL-1K)~\cite{albanie2020bsl} has not been released to the public yet at the moment of writing. A statistical summary of the mentioned datasets is provided in Table~\ref{dataset_table}.

\subsection{Multi-modal Data Preparation}

\noindent\textbf{Whole-body Pose Keypoints and Features}. 
MMPose~\cite{mmpose2020} provides a whole-body pose estimator pretrained with whole-body keypoint annotations and HRNet~\cite{sun2019deep} as its backbone. We use the pretrained model to obtain 133 keypoints from RGB videos, based on which we construct the 27-node 2D graph in Section~\ref{sec:graph_construction} and process the 3D graphs using depth videos. 
Keypoint coordinates are normalized to [-1,1]. Random sampling, mirroring, rotating, scaling, jittering, and shifting are applied as data augmentations. We use a sample duration of 150 frames. We repeat videos with lesser than 150 frames to 150 frames. To obtain skeleton features for each video, we downsample the estimated heatmaps, choose 33 joint channels, and uniformly sample $60$ frames.

\noindent\textbf{RGB Frames and Optical Flow}. 
We extract all frames from RGB videos to load and process them faster in parallel. 
Besides, we use the Denseflow API implemented with OpenCV and CUDA that provided by OpenMMLab~\cite{denseflow} to obtain the optical flows using the TVL1 algorithm~\cite{zach2007duality}. The output x and y flow maps are concatenated in the channel dimension. We cropped all RGB and optical flow frames to the bounding boxes from the pose estimator, and then resize them to $256 \times 256$. When we train the model, we sample 32 consecutive frames randomly from the input video. When testing, we sample five such clips uniformly from the input videos and average on their predicted scores. 

\begin{table}[!t]
\caption{Performance of multi-stream SL-GCN.}
\label{tab:sl-gcn_results}
\centering
\begin{tabular}{l | c  c | c  c | c  c}
\hline
\multirow{2}{*}{Streams} & \multicolumn{2}{c}{AUTSL} & \multicolumn{2}{|c}{SLR500} & \multicolumn{2}{|c}{WLASL2000}\\
 & Top-1 & Top-5 & Top-1 & Top-5 & Top-1 & Top-5 \\
\hline
Joint       & 95.35 & 99.49 & 97.90 & 99.92 & 45.61 & 77.79\\
Bone        & 95.69 & 99.55 & 97.93 & 99.88 & 43.27 & 75.58\\
Joint Motion& 93.21 & 99.12 & 97.04 & 99.80 & 27.23 & 56.73\\
Bone Motion & 93.29 & 99.31 & 97.24 & 99.81 & 31.26 & 60.35\\
\hline
Multi-stream &\textbf{96.47} & \textbf{99.76} &\textbf{98.16} & \textbf{99.95} &\textbf{51.50} & \textbf{84.94}\\
\hline
\end{tabular}
\end{table}

\begin{table}[!t]
\caption{Ablation studies on SL-GCN on AUTSL validation set.}
\label{tab:ablation_gcn}
\centering
\begin{tabular}{l | c}
\hline
Variations & Top-1 \\
\hline
SL-GCN (Joint) & \textbf{95.02} \\
w/o Graph Reduction & 63.69\\
w/o Decouple GCN & 94.66\\
w/o Drop Graph & 94.81\\
w/o Keypoints Augmentation & 90.16\\
w/o STC Attention & 93.53\\
\hline
\end{tabular}
\end{table}

\noindent\textbf{Depth HHA and Depth Flow}. 
HHA stands for the horizontal disparity, height above the ground, and angle normal that encode a depth map into a three-channel RGB image. 
Because HHA features enable better scene understanding, we extract HHA features from depth videos as another modality instead of inputting gray-scale depth maps directly. The depth videos of the AUTSL dataset come with masks, so we mask out those regions and fill them with zeros when generating HHA. An HHA example is shown in Figure~\ref{fig:modalities}(c), where black regions are masked out pixels. Cropping and resizing operations are performed the same way as RGB frames. We also apply the same data augmentations to HHA features as the RGB modality. 
Apart from that, we use Denseflow to extract optical flow from the depth modality as well. A sample of depth flow is shown in Figure~\ref{fig:modalities}(e). Compared with optical flows (\eg Figure~\ref{fig:modalities}(d)), depth flows contain less noise and capture distinctive motion information.

\begin{table}[!t]
\caption{Results of single modalities on AUTSL test set.}
\label{tab:autsl_test_single}
\centering
\begin{tabular}{l | c  c}
\hline
Modality & Top-1 & Top-5\\
\hline
Keypoints 2D    & 96.47 & 99.76\\
Keypoints 3D    & \textbf{96.53} & \textbf{99.81}\\
Features        & 93.37 & 99.20\\
RGB Frames      & 95.00 & 99.47\\
RGB Flow        & 90.41 & 98.74\\
Depth HHA       & 93.75 & 99.28\\
Depth Flow      & 90.78 & 98.50\\
\hline
\end{tabular}
\end{table}

\subsection{Performance of Multi-stream SL-GCN}
Both top-1 and top-5 recognition rates of the proposed multi-stream SL-GCN are reported in Table~\ref{tab:sl-gcn_results}. Among the four streams, the joint stream provides the best accuracy. The multi-stream approach further improves the overall accuracy, which demonstrates that our proposed whole-body skeleton graph and multi-stream SL-GCN are very effective. Table~\ref{tab:autsl_test_single} shows that SL-GCN (Keypoints 2D and 3D) perform the best in all single-modality methods. Since the skeleton graphs are less complicated, the graph-based method is lighter-weight and much faster to inference compared with large-capacity 3DCNN-based models, which is another advantage of the proposed SL-GCN.

Table~\ref{tab:ablation_gcn} presents ablation studies on the proposed SL-GCN. We find that the graph reduction technique contributes the most to the recognition rate. Without that, it can barely learn from the noisy dynamics in the 133-node skeleton graph. Due to the limitation of annotated data, the model is easy to overfit, thus the data augmentation techniques are also important in learning the embedded dynamics. Besides, we find that the DropGraph module, the decoupling GCN module, and the STC attention mechanism all contribute to the final performance.

\subsection{Multi-modal Performance on AUTSL Dataset}
The results of all single-modal methods on the AUTSL balanced test set are reported in Table~\ref{tab:autsl_test_single}. The Keypoints 2D and Keypoints 3D method represents our proposed multi-stream SL-GCN using 2D and 3D skeleton graphs. They perform the best compared with the other methods. 
We also find that the depth flow achieves a little better recognition rate compared with the RGB flow resulted from lesser noisy data.  
The fused ensembles in both RGB and RGB-D scenarios using different choices of modalities are summarized in Table~\ref{tab:multi-modal-autsl} as three groups. The skeleton-based ensemble (Skel 2D and Skel 3D) achieves better accuracy than ``RGB + Flow'' and Depth ensemble (HHA + Depth Flow), which demonstrates that the proposed skeleton-based methods are very effective in SLR. The RGB and RGB-D ensemble results that use all available modalities show that the skeleton-based methods also complement RGB and depth modalities. Their collaboration further improves the overall performance. 
Comparisons with other players in the challenge are reported in Table~\ref{tab:comparison_autsl}. Our improved SAM-SLR-v2 achieves better recognition rates compared with its previous version SAM-SLR that won the championships of the SLR challenge. During the challenge, we are allowed to use the validation labels to fine-tune our models. Our results with fine-tuning on the validation set are shown at the bottom. Our results without fine-tuning still surpass the other methods fine-tuned with validation labels. Our method achieves the state-of-the-art with significant margins in both RGB and RGB-D scenarios. 

\begin{table}[!t]
\caption{Multi-modal ensemble results evaluated on AUTSL test set. Abbrevs: K2=Keypoints 2D; K3=Keypoints 3D; F=Features; R=RGB; O=Optical Flow; H=HHA; D=Depth Flow.}
\label{tab:multi-modal-autsl}
\centering
\begin{tabular}{l | c | c | c | c | c | c | c | c  c}
\hline
Ensemble & K2 & K3 & F & R & O & H & D & Top-1 & Top-5\\
\hline
Skel 2D  & \checkmark && \checkmark &&&&&  96.90 & 99.84  \\
RGB+Flow &&& \checkmark && \checkmark &&& 96.10  & 99.73 \\
RGB All & \checkmark && \checkmark & \checkmark & \checkmark &&& \textbf{97.62} & \textbf{100}  \\
\hline
Skel 3D  && \checkmark & \checkmark &&&&& 97.01  & 99.87  \\
Depth &&&&&& \checkmark & \checkmark & 95.38  &  99.57\\
RGB\&D &&& & \checkmark & \checkmark & \checkmark & \checkmark & 96.73 & 99.73 \\
RGBD All && \checkmark & \checkmark & \checkmark & \checkmark & \checkmark & \checkmark & \textbf{98.02} & \textbf{100} \\
\hline
\end{tabular}
\end{table}

\begin{table}[t]
\caption{Performance our ensemble results (with and without fine-tuning using validation set) evaluated on AUTSL test set.}
\label{tab:comparison_autsl}
\centering
\begin{tabular}{l | c | c  c | c  c }
\hline
\multirow{2}{*}{Model} & \multirow{2}{*}{Fine-tune} & \multicolumn{2}{c}{RGB} &  \multicolumn{2}{|c}{RGB-D}\\
  &  & Top-1 & Top-5 & Top-1 & Top-5 \\
\hline
Baseline~\cite{SLR_dataset} & - & 49.22 & 75.78 & 62.02 & 83.45\\
VTN-PF~\cite{de2021isolated} & w/ val & 92.92 & - & 93.32 & -\\
wenbinwuee & w/ val &96.55 & - & 96.55 & -\\
USTC-SLR & w/ val & 97.62 & - & 97.65 & -\\
\hline
SAM-SLR~\cite{jiang2021skeleton} & No & 97.51 & \textbf{100} & 97.68 & \textbf{100}\\
SAM-SLR-v2 & No & \textbf{98.00} & \textbf{100} & \textbf{98.10} & \textbf{100}\\
w/o Keypoint 3D & No & - & - & 98.02 & 99.95\\
\hline
+ Extra data & w/ val & \textbf{98.42} & \textbf{100} & \textbf{98.53} & \textbf{100}\\
\hline
\end{tabular}
\end{table}

\subsection{Multi-modal Performance on SLR500 Dataset}
Performance on the isolated Chinese sign language dataset (SLR500) is reported in Table~\ref{tab:validation_results_all_csl}. Since this dataset is collected in a lab-controlled environment with a large number of repetitions and invariant background, all modalities obtain higher recognition rates compared with the other two datasets. For the same reason, compared with skeleton-based methods, RGB-based 3DCNN achieves a higher Top-1 recognition rate but a lower Top-5 recognition rate. Since the SLR500 dataset provides only a validation set instead of train-val-test splits, we use the weights learned from the AUTSL dataset to obtain the final ensemble results. Compared with the recent state-of-the-art method GLE-Net~\cite{hu2020global} and Hand+RGB~\cite{hu2021hand}, our two-modality methods and all-modality SAM-SLR-v2 all achieve much better recognition rates.

\begin{table}[t]
\caption{Multi-modal performance on SLR500 dataset.}
\label{tab:validation_results_all_csl}
\centering
\begin{tabular}{l | c c}
\hline
Modality & Top-1 & Top-5 \\
\hline
GLE-Net~\cite{hu2020global} & 96.80 & -\\
Hand + RGB~\cite{hu2021hand} & 98.30 & -\\

\hline
Keypoints       & 98.16 & \textbf{99.95}\\
Features        & 97.34 & 99.80\\
RGB Frames      & \textbf{98.26} & 99.84\\
RGB Flow        & 95.94 & 99.63\\
\hline
Key + Feat      & 98.56 & 99.96\\
Key + RGB       & \textbf{98.98} & \textbf{99.97}\\
RGB + Flow      & 98.45 & 99.88\\
\hline
SAM-SLR~\cite{jiang2021skeleton} & 98.98 & \textbf{99.98}\\
SAM-SLR-v2 & \textbf{99.00} & \textbf{99.98}\\
\hline
\end{tabular}
\end{table}

\subsection{Multi-modal Performance on WLASL2000}
The WLASL2000 dataset is the most challenging variation of the WLASL dataset with unconstrained backgrounds and camera conditions. Since the dataset is not balanced, we report both per-instance accuracy as well as per-class accuracy following \cite{vaezijoze2019ms-asl,hu2021hand}. As reported in Table~\ref{tab:validation_results_all_wlasl}, the keypoints modality performs the best (51.50\% per-instance) among all the single-modal methods due to its independence on backgrounds and resistance to noises. The keypoints and RGB modalities complement each other and boost the two-modal performance. Using all the available modalities, the top-1 performance is improved to 57.55\% per-instance. Our proposed late ensemble model is capable of raising the recognition rates by around 0.7\%. Compared with other state-of-the-art methods, our proposed SAM-SLR-v2 achieves the best recognition rates in both per-instance and per-class metrics with large leading gaps of around 8\%.

\begin{table}[t]
\caption{Multi-modal performance on WLASL2000 dataset. \\ * pretrained with extra BSL-1K dataset which is not publicly available at the moment.}
\label{tab:validation_results_all_wlasl}
\centering
\begin{tabular}{l | c  c | c  c}
\hline
\multirow{2}{*}{Modality} & \multicolumn{2}{|c}{per-instance} &  \multicolumn{2}{|c}{per-class}\\
 & Top-1 & Top-5 & Top-1 & Top-5 \\
\hline
Baseline~\cite{li2020word} & 32.48 & 57.31 & - & -\\
Fusion-3~\cite{hosain2021hand} & 38.84 & 67.58 & - & -\\
I3D~\cite{albanie2020bsl} * & 46.82 & 79.36 & 44.72 & 78.47\\
Hand + RGB~\cite{hu2021hand} & 51.39 & 86.34 & 48.75 & 85.74\\
\hline
Keypoints       & \textbf{51.50} & \textbf{84.94} & \textbf{48.87} & \textbf{84.02}\\
Features        & 46.84 & 79.63 & 44.41 & 78.35\\
RGB Frames      & 47.51 & 80.31 & 44.53 & 78.93\\
RGB Flow        & 40.46 & 73.23 & 37.88 & 71.86\\
\hline
Key + Feat      & 55.03 & 89.90 & 52.48 & 87.03\\
Key + RGB       & \textbf{57.55} & \textbf{90.34} & \textbf{54.83} & \textbf{89.75}\\
RGB + Flow      & 53.53 & 86.50 & 50.63 & 85.77\\
\hline
SAM-SLR~\cite{jiang2021skeleton}    & 58.73 & 91.46 & 55.93 & \textbf{90.94}\\
SAM-SLR-v2 & \textbf{59.39} & \textbf{91.48} & \textbf{56.63} & 90.89\\
\hline
\end{tabular}
\end{table}

\begin{figure*}[!t]
  \centering
  \includegraphics[width=0.99\textwidth]{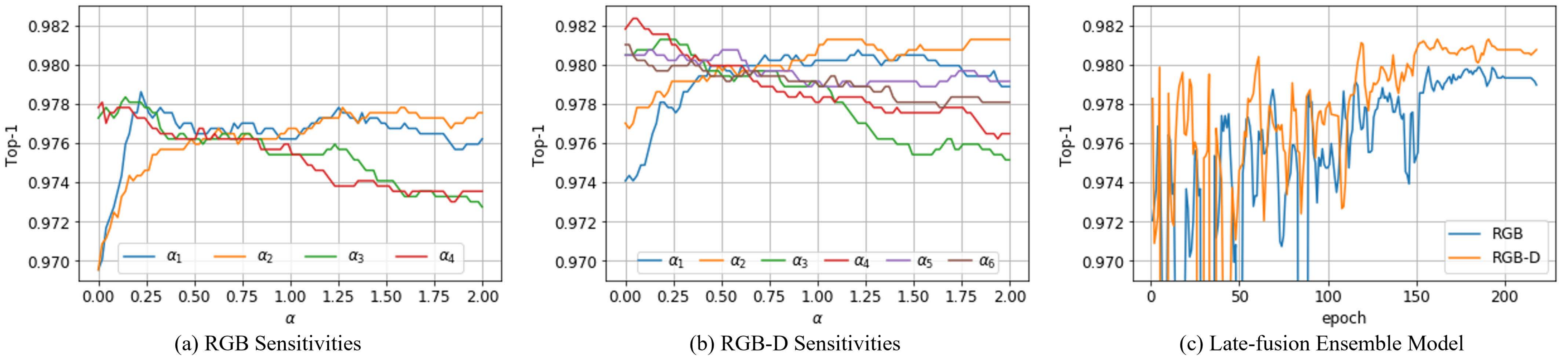}
  \caption{Sensitivity analysis of ensemble models evaluated using AUTSL test set. (a) Sensitivities of four RGB ensemble parameters ($\alpha_1\dots\alpha_4$); (b) Sensitivities of on six RGB-D ensemble parameters ($\alpha_1\dots\alpha_6$); (c) Learning the ensemble weights using the proposed late-fusion GEM (both RGB and RGB-D).}
  \label{fig:sensitivity}
\end{figure*}

\subsection{Ablation Study on SSTCN and 3DCNN}
\label{sec:ablation study SSTCN}
An ablation study on SSTCN is provided in Table~\ref{tab:ablation_sstcn}. Compared with ResNet3D~\cite{hara2018can,kataoka2020would} and ResNet2+1D~\cite{tran2018closer} on the same feature size, SSTCN is more effective on recognizing sign glosses using skeleton features. 
Our study also reveals that the proposed SSTCN can achieve even higher accuracy with larger input feature sizes. 

Table~\ref{tab:ablation_rgb} shows an ablation study of 3D CNN architecture using the RGB frames. It shows that both swish activation and label smoothing are effective, where they improve the top-1 accuracy by 2\% and 1\%, respectively. Pretraining on the CSL dataset~\cite{zhang2016chinese} improves the overall accuracy by 1.4\%. The ResNet2+1D-18 backbone provides 1.7\% better performance than ResNet3D-18.

\begin{table}[!t]
\caption{Comparing SSTCN with ResNet3D and ResNet2+1D and ablation study on feature sizes evaluated on AUTSL validation set.}
\label{tab:ablation_sstcn}
\centering
\begin{tabular}{l | c | c}
\hline
Methods & Feature size & Top-1 \\
\hline
ResNet3D       & $12\times12$ & 92.82\\
ResNet2+1D     & $12\times12$ & 93.03\\
\hline
SSTCN & $12\times12$ & 93.60\\
SSTCN &$24\times24$ & \textbf{94.32}\\
\hline
\end{tabular}
\end{table}

\subsection{Comparison between Multi-modal Ensembles}
The model-free simple ensemble method requires a lot of effort in tuning the weights of different modalities. With our proposed late-fusion GEM, those weights are automatically learned from the data. We study the effectiveness of GEM compared with the simple ensemble method. In Table~\ref{tab:comparison_autsl}, \ref{tab:validation_results_all_csl}, and \ref{tab:validation_results_all_wlasl}, performance of model-free simple ensemble is 
shown as SAM-SLR. Compared with the simple ensemble, our proposed late-fusion GEM can improve the overall performance as well as avoid hard tuning on modality weights.

\subsection{Sensitivity Analysis of Ensemble Methods}
A sensitivity study on ensemble methods is conducted on the AUTSL test dataset. In the experiment, we use the manually tuned parameters in Section~\ref{sec:ensemble} as our basis and change one parameter at a time (varying from 0.0 to 2.0), while keeping the other parameters fixed. We plot the resulted top-1 accuracy to analyze the sensitivity of those ensemble parameters. In the RGB scenario, we have four modalities and we analyze the sensitivity of $\alpha_1$ to $\alpha_4$, as shown in Figure~\ref{fig:sensitivity}(a). In the RGB-D scenario, we analyze the sensitivity of $\alpha_1$ to $\alpha_6$, as shown in Figure~\ref{fig:sensitivity}(b). 

\begin{table}[!t]
\caption{An ablation study of 3D CNN architecture using RGB frames evaluated on AUTSL validation set.}
\label{tab:ablation_rgb}
\centering
\begin{tabular}{l | c}
\hline
3D CNN Variations & Top-1 \\
\hline
Ours (RGB Frame) & \textbf{94.77} \\
w/o Label Smoothing & 93.75\\
w/o Swish Activation & 92.88\\
w/o Pretraining on CSL & 93.41\\
w/ ResNet3D-18 Backbone & 93.10\\
\hline
\end{tabular}
\end{table}

In this experiment, the ensemble parameters tuned on the validation set do not provide the highest performance on the test set. The ensemble performance is sensitive to the values of the parameters that small changes may lead to large variations in the ensemble accuracy. Moreover, when there are more modalities (\eg in the RGB-D scenario), it is hard to optimize all those parameters manually. Our proposed late-fusion GEM solves this problem by automatically learning these parameters from data. Figure~\ref{fig:sensitivity}(c) shows the top-1 accuracy during training the proposed model. Initially, the ensemble model gives sensitive results due to its large search space. The model converges after 100 epochs and delivers steady and accurate ensemble results. In summary, the proposed late-fusion GEM is very effective in learning the multi-modal ensemble, gives robust higher performance than simple ensemble, and saves a lot of effort to tune the best parameters.


\subsection{Challenging Cases and Model Limitations}
\begin{figure}[t]
  \centering
  \includegraphics[width=0.47\textwidth]{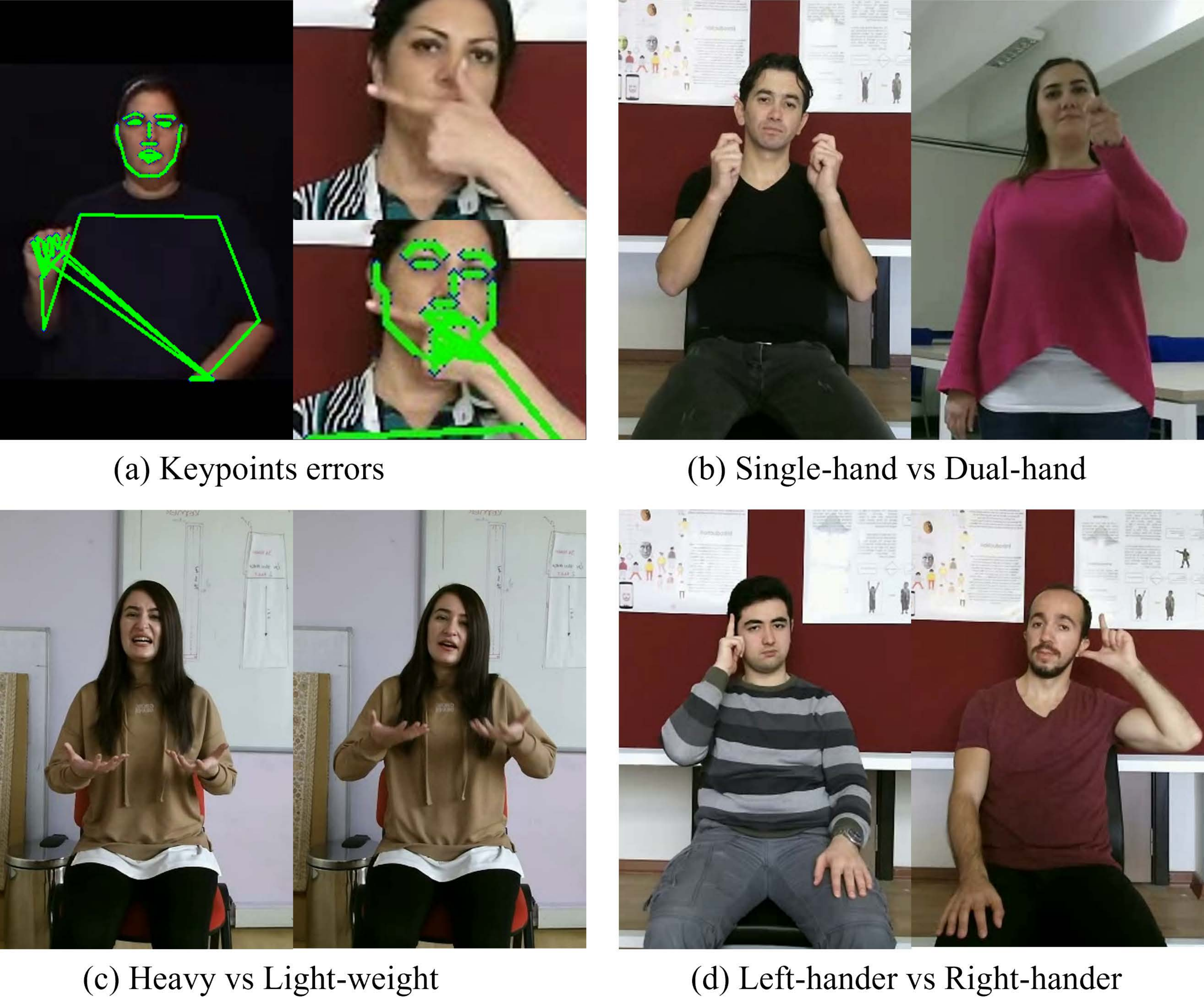}
  \caption{Examples of challenging cases: (a) Errors in whole-body keypoint estimation; (b) A same sign performed differently by different signers; (c) Two opposite signs can be visually similar; (d) Left-handers and right-handers perform mirror-symmetrically.}
  \label{fig:case_analysis}
\end{figure}

Figure~\ref{fig:case_analysis} shows some challenging cases of sign language recognition from the AUTSL dataset. The offline full-body pose estimator may fail due to off-screen or occlusion, especially for fingers, which is a common issue for pose estimation methods. However, fingers play a critical role in expressing signs. Some of those failures can be corrected by RGB-based features. That is the reason why multi-modal ensemble significantly boosts the recognition rate. Moreover, a same sign may be performed very differently by different signers (\eg signers may use their left hands, right hands, or both hands to perform the same gloss). Hence, mirror augmentation is very important in model training, but more data collected from more signers are desired. Last, opposite signs could be visually similar in a single frame (\eg heavy vs light-weight). Distinguishing between those signs requires delicate modeling of spatio-temporal dynamics. In our experiment, we find that skeleton-based methods are smarter choices over RGB-based methods.

%% file: main.bbl
\begin{thebibliography}{10}
\providecommand{\url}[1]{#1}
\csname url@samestyle\endcsname
\providecommand{\newblock}{\relax}
\providecommand{\bibinfo}[2]{#2}
\providecommand{\BIBentrySTDinterwordspacing}{\spaceskip=0pt\relax}
\providecommand{\BIBentryALTinterwordstretchfactor}{4}
\providecommand{\BIBentryALTinterwordspacing}{\spaceskip=\fontdimen2\font plus
\BIBentryALTinterwordstretchfactor\fontdimen3\font minus
  \fontdimen4\font\relax}
\providecommand{\BIBforeignlanguage}[2]{{%
\expandafter\ifx\csname l@#1\endcsname\relax
\typeout{** WARNING: IEEEtran.bst: No hyphenation pattern has been}%
\typeout{** loaded for the language `#1'. Using the pattern for}%
\typeout{** the default language instead.}%
\else
\language=\csname l@#1\endcsname
\fi
#2}}
\providecommand{\BIBdecl}{\relax}
\BIBdecl

\bibitem{SL_intro}
K.~Emmorey, \emph{Language, cognition, and the brain: Insights from sign
  language research}.\hskip 1em plus 0.5em minus 0.4em\relax Psychology Press,
  2001.

\bibitem{SL_book1}
C.~Valli and C.~Lucas, \emph{Linguistics of American sign language: An
  introduction}.\hskip 1em plus 0.5em minus 0.4em\relax Gallaudet University
  Press, 2000.

\bibitem{SL_book2}
T.~Johnston and A.~Schembri, \emph{Australian Sign Language (Auslan): An
  introduction to sign language linguistics}.\hskip 1em plus 0.5em minus
  0.4em\relax Cambridge University Press, 2007.

\bibitem{yang2010chinese}
Q.~Yang, ``Chinese sign language recognition based on video sequence appearance
  modeling,'' in \emph{Proceedings of IEEE Conference on Industrial Electronics
  and Applications}, 2010, pp. 1537--1542.

\bibitem{different_culture}
A.~Mindess, \emph{Reading between the signs: Intercultural communication for
  sign language interpreters}.\hskip 1em plus 0.5em minus 0.4em\relax Nicholas
  Brealey, 2014.

\bibitem{hog_feature}
Q.~Zhu, M.-C. Yeh, K.-T. Cheng, and S.~Avidan, ``Fast human detection using a
  cascade of histograms of oriented gradients,'' in \emph{Proceedings of IEEE
  Computer Vision and Pattern Recognition}, vol.~2, 2006, pp. 1491--1498.

\bibitem{sift_feature}
D.~G. Lowe, ``Object recognition from local scale-invariant features,'' in
  \emph{Proceedings of IEEE International Conference on Computer Vision},
  vol.~2, 1999, pp. 1150--1157.

\bibitem{dardas2011real}
N.~H. Dardas and N.~D. Georganas, ``Real-time hand gesture detection and
  recognition using bag-of-features and support vector machine techniques,''
  \emph{IEEE Transactions on Instrumentation and measurement}, vol.~60, no.~11,
  pp. 3592--3607, 2011.

\bibitem{memics2013kinect}
A.~Memi{\c{s}} and S.~Albayrak, ``A {K}inect based sign language recognition
  system using spatio-temporal features,'' in \emph{Proceedings of
  International Conference on Machine Vision}, vol. 9067.\hskip 1em plus 0.5em
  minus 0.4em\relax International Society for Optics and Photonics, 2013, p.
  90670X.

\bibitem{sincan2019isolated}
O.~M. Sincan, A.~O. Tur, and H.~Y. Keles, ``Isolated sign language recognition
  with multi-scale features using {LSTM},'' in \emph{2019 27th Signal
  Processing and Communications Applications Conference (SIU)}.\hskip 1em plus
  0.5em minus 0.4em\relax IEEE, 2019, pp. 1--4.

\bibitem{li2020word}
D.~Li, C.~Rodriguez, X.~Yu, and H.~Li, ``Word-level deep sign language
  recognition from video: {A} new large-scale dataset and methods comparison,''
  in \emph{Proceedings of IEEE Winter Conference on Applications of Computer
  Vision}, 2020, pp. 1459--1469.

\bibitem{pigou2018beyond}
L.~Pigou, A.~Van Den~Oord, S.~Dieleman, M.~Van~Herreweghe, and J.~Dambre,
  ``Beyond temporal pooling: {R}ecurrence and temporal convolutions for gesture
  recognition in video,'' \emph{International Journal of Computer Vision}, vol.
  126, no.~2, pp. 430--439, 2018.

\bibitem{tur2019isolated}
A.~O. Tur and H.~Y. Keles, ``Isolated sign recognition with a siamese neural
  network of {RGB} and depth streams,'' in \emph{IEEE International Conference
  on Smart Technologies}, 2019, pp. 1--6.

\bibitem{huang2018video}
J.~Huang, W.~Zhou, Q.~Zhang, H.~Li, and W.~Li, ``Video-based sign language
  recognition without temporal segmentation,'' in \emph{Proceedings of AAAI
  Conference on Artificial Intelligence}, vol.~32, no.~1, 2018.

\bibitem{huang2018attention}
J.~Huang, W.~Zhou, H.~Li, and W.~Li, ``Attention-based {3D-CNN}s for
  large-vocabulary sign language recognition,'' \emph{IEEE Transactions on
  Circuits and Systems for Video Technology}, vol.~29, no.~9, pp. 2822--2832,
  2018.

\bibitem{hu2021hand}
H.~Hu, W.~Zhou, and H.~Li, ``Hand-model-aware sign language recognition,'' in
  \emph{Proceedings of the AAAI Conference on Artificial Intelligence},
  vol.~35, no.~2, 2021, pp. 1558--1566.

\bibitem{hosain2021hand}
A.~A. Hosain, P.~S. Santhalingam, P.~Pathak, H.~Rangwala, and J.~Kosecka,
  ``Hand pose guided 3d pooling for word-level sign language recognition,'' in
  \emph{Proceedings of the IEEE/CVF Winter Conference on Applications of
  Computer Vision}, 2021, pp. 3429--3439.

\bibitem{zhou2021spatial}
H.~Zhou, W.~Zhou, Y.~Zhou, and H.~Li, ``Spatial-temporal multi-cue network for
  sign language recognition and translation,'' \emph{IEEE Transactions on
  Multimedia}, 2021.

\bibitem{yan2018spatial}
S.~Yan, Y.~Xiong, and D.~Lin, ``Spatial temporal graph convolutional networks
  for skeleton-based action recognition,'' in \emph{Proceedings of AAAI
  conference on Artificial Intelligence}, vol.~32, no.~1, 2018.

\bibitem{skeleton_tsrnn}
H.~Wang and L.~Wang, ``Modeling temporal dynamics and spatial configurations of
  actions using two-stream recurrent neural networks,'' in \emph{Proceedings of
  IEEE Computer Vision and Pattern Recognition}, 2017.

\bibitem{shi2020skeleton}
L.~Shi, Y.~Zhang, J.~Cheng, and H.~Lu, ``Skeleton-based action recognition with
  multi-stream adaptive graph convolutional networks,'' \emph{IEEE Transactions
  on Image Processing}, vol.~29, pp. 9532--9545, 2020.

\bibitem{chengdecoupling}
K.~Cheng, Y.~Zhang, C.~Cao, L.~Shi, J.~Cheng, and H.~Lu, ``Decoupling {GCN}
  with {DropGraph} module for skeleton-based action recognition,'' in
  \emph{Proceedings of European Conference on Computer Vision}, 2020.

\bibitem{song2020stronger}
Y.-F. Song, Z.~Zhang, C.~Shan, and L.~Wang, ``Stronger, faster and more
  explainable: {A} graph convolutional baseline for skeleton-based action
  recognition,'' in \emph{Proceedings of ACM International Conference on
  Multimedia}, 2020, pp. 1625--1633.

\bibitem{xiao2020skeleton}
Q.~Xiao, M.~Qin, and Y.~Yin, ``Skeleton-based chinese sign language recognition
  and generation for bidirectional communication between deaf and hearing
  people,'' \emph{Neural Networks}, vol. 125, pp. 41--55, 2020.

\bibitem{jiang2021skeleton}
S.~Jiang, B.~Sun, L.~Wang, Y.~Bai, K.~Li, and Y.~Fu, ``Skeleton aware
  multi-modal sign language recognition,'' in \emph{Proceedings of the IEEE/CVF
  Conference on Computer Vision and Pattern Recognition}, 2021, pp. 3413--3423.

\bibitem{Sincan:CVPRW:2021}
O.~M. Sincan, J.~C.~S. Jacques~Junior, S.~Escalera, and H.~Y. Keles, ``Chalearn
  {LAP} large scale signer independent isolated sign language recognition
  challenge: Design, results and future research,'' in \emph{Proceedings of the
  IEEE/CVF Conference on Computer Vision and Pattern Recognition Workshops},
  2021.

\bibitem{lim2019isolated}
K.~M. Lim, A.~W.~C. Tan, C.~P. Lee, and S.~C. Tan, ``Isolated sign language
  recognition using convolutional neural network hand modelling and hand energy
  image,'' \emph{Multimedia Tools and Applications}, vol.~78, no.~14, pp.
  19\,917--19\,944, 2019.

\bibitem{koller2018deep}
O.~Koller, S.~Zargaran, H.~Ney, and R.~Bowden, ``Deep sign: {E}nabling robust
  statistical continuous sign language recognition via hybrid {CNN-HMM}s,''
  \emph{International Journal of Computer Vision}, vol. 126, no.~12, pp.
  1311--1325, 2018.

\bibitem{guo2018hierarchical}
D.~Guo, W.~Zhou, H.~Li, and M.~Wang, ``Hierarchical {LSTM} for sign language
  translation,'' in \emph{Proceedings of AAAI Conference on Artificial
  Intelligence}, vol.~32, no.~1, 2018.

\bibitem{pu2019iterative}
J.~Pu, W.~Zhou, and H.~Li, ``Iterative alignment network for continuous sign
  language recognition,'' in \emph{Proceedings of IEEE Conference on Computer
  Vision and Pattern Recognition}, 2019, pp. 4165--4174.

\bibitem{cui2019deep}
R.~Cui, H.~Liu, and C.~Zhang, ``A deep neural framework for continuous sign
  language recognition by iterative training,'' \emph{IEEE Transactions on
  Multimedia}, vol.~21, no.~7, pp. 1880--1891, 2019.

\bibitem{cai2021jolo}
J.~Cai, N.~Jiang, X.~Han, K.~Jia, and J.~Lu, ``Jolo-gcn: Mining joint-centered
  light-weight information for skeleton-based action recognition,'' in
  \emph{Proceedings of the IEEE/CVF Winter Conference on Applications of
  Computer Vision}, 2021, pp. 2735--2744.

\bibitem{du2015hierarchical}
Y.~Du, W.~Wang, and L.~Wang, ``Hierarchical recurrent neural network for
  skeleton based action recognition,'' in \emph{Proceedings of the IEEE
  conference on computer vision and pattern recognition}, 2015, pp. 1110--1118.

\bibitem{huang2017deep}
Z.~Huang, C.~Wan, T.~Probst, and L.~Van~Gool, ``Deep learning on lie groups for
  skeleton-based action recognition,'' in \emph{Proceedings of the IEEE
  conference on computer vision and pattern recognition}, 2017, pp. 6099--6108.

\bibitem{li2019actional}
M.~Li, S.~Chen, X.~Chen, Y.~Zhang, Y.~Wang, and Q.~Tian, ``Actional-structural
  graph convolutional networks for skeleton-based action recognition,'' in
  \emph{Proceedings of the IEEE/CVF Conference on Computer Vision and Pattern
  Recognition}, 2019, pp. 3595--3603.

\bibitem{li2018independently}
S.~Li, W.~Li, C.~Cook, C.~Zhu, and Y.~Gao, ``Independently recurrent neural
  network (indrnn): Building a longer and deeper rnn,'' in \emph{Proceedings of
  the IEEE conference on computer vision and pattern recognition}, 2018, pp.
  5457--5466.

\bibitem{liu2016spatio}
J.~Liu, A.~Shahroudy, D.~Xu, and G.~Wang, ``Spatio-temporal lstm with trust
  gates for 3d human action recognition,'' in \emph{European conference on
  computer vision}.\hskip 1em plus 0.5em minus 0.4em\relax Springer, 2016, pp.
  816--833.

\bibitem{baradel2017human}
F.~Baradel, C.~Wolf, and J.~Mille, ``Human action recognition: Pose-based
  attention draws focus to hands,'' in \emph{Proceedings of the IEEE
  International Conference on Computer Vision Workshops}, 2017, pp. 604--613.

\bibitem{carreira2017quo}
J.~Carreira and A.~Zisserman, ``Quo vadis, action recognition? a new model and
  the kinetics dataset,'' in \emph{proceedings of the IEEE Conference on
  Computer Vision and Pattern Recognition}, 2017, pp. 6299--6308.

\bibitem{choutas2018potion}
V.~Choutas, P.~Weinzaepfel, J.~Revaud, and C.~Schmid, ``Potion: Pose motion
  representation for action recognition,'' in \emph{Proceedings of the IEEE
  conference on computer vision and pattern recognition}, 2018, pp. 7024--7033.

\bibitem{hu2018deep}
J.-F. Hu, W.-S. Zheng, J.~Pan, J.~Lai, and J.~Zhang, ``Deep bilinear learning
  for rgb-d action recognition,'' in \emph{Proceedings of the European
  Conference on Computer Vision}, 2018, pp. 335--351.

\bibitem{zolfaghari2017chained}
M.~Zolfaghari, G.~L. Oliveira, N.~Sedaghat, and T.~Brox, ``Chained multi-stream
  networks exploiting pose, motion, and appearance for action classification
  and detection,'' in \emph{Proceedings of the IEEE International Conference on
  Computer Vision}, 2017, pp. 2904--2913.

\bibitem{si2018skeleton}
C.~Si, Y.~Jing, W.~Wang, L.~Wang, and T.~Tan, ``Skeleton-based action
  recognition with spatial reasoning and temporal stack learning,'' in
  \emph{Proceedings of the European Conference on Computer Vision}, 2018, pp.
  103--118.

\bibitem{shi2019skeleton}
L.~Shi, Y.~Zhang, J.~Cheng, and H.~Lu, ``Skeleton-based action recognition with
  directed graph neural networks,'' in \emph{Proceedings of the IEEE/CVF
  Conference on Computer Vision and Pattern Recognition}, 2019, pp. 7912--7921.

\bibitem{shi2019two}
------, ``Two-stream adaptive graph convolutional networks for skeleton-based
  action recognition,'' in \emph{Proceedings of the IEEE/CVF Conference on
  Computer Vision and Pattern Recognition}, 2019, pp. 12\,026--12\,035.

\bibitem{si2019attention}
C.~Si, W.~Chen, W.~Wang, L.~Wang, and T.~Tan, ``An attention enhanced graph
  convolutional lstm network for skeleton-based action recognition,'' in
  \emph{Proceedings of the IEEE/CVF Conference on Computer Vision and Pattern
  Recognition}, 2019, pp. 1227--1236.

\bibitem{he2016deep}
K.~He, X.~Zhang, S.~Ren, and J.~Sun, ``Deep residual learning for image
  recognition,'' in \emph{Proceedings of the IEEE conference on computer vision
  and pattern recognition}, 2016, pp. 770--778.

\bibitem{de2019spatial}
C.~C. de~Amorim, D.~Mac{\^e}do, and C.~Zanchettin, ``Spatial-temporal graph
  convolutional networks for sign language recognition,'' in
  \emph{International Conference on Artificial Neural Networks}.\hskip 1em plus
  0.5em minus 0.4em\relax Springer, 2019, pp. 646--657.

\bibitem{zheng2016cross}
J.~Zheng, Z.~Jiang, and R.~Chellappa, ``Cross-view action recognition via
  transferable dictionary learning,'' \emph{IEEE Transactions on Image
  Processing}, vol.~25, no.~6, pp. 2542--2556, 2016.

\bibitem{wang2013dense}
H.~Wang, A.~Kl{\"a}ser, C.~Schmid, and C.-L. Liu, ``Dense trajectories and
  motion boundary descriptors for action recognition,'' \emph{International
  Journal of Computer Vision}, vol. 103, no.~1, pp. 60--79, 2013.

\bibitem{hoff_CVPR16}
J.~Hoffman, S.~Gupta, and T.~Darrell, ``Learning with side information through
  modality hallucination,'' in \emph{Proceedings of IEEE Computer Vision and
  Pattern Recognition}, 2016, pp. 826--834.

\bibitem{Wang_2018_ECCV}
D.~Wang, W.~Ouyang, W.~Li, and D.~Xu, ``Dividing and aggregating network for
  multi-view action recognition,'' in \emph{Proceedings of European Conference
  on Computer Vision}, September 2018.

\bibitem{pm_gan}
L.~Wang, C.~Gao, L.~Yang, Y.~Zhao, W.~Zuo, and D.~Meng, ``{PM-GAN}s:
  {D}iscriminative representation learning for action recognition using
  partial-modalities,'' in \emph{Proceedings of European Conference on Computer
  Vision}, 2018, pp. 384--401.

\bibitem{shahroudy2018deep}
A.~Shahroudy, T.-T. Ng, Y.~Gong, and G.~Wang, ``Deep multimodal feature
  analysis for action recognition in {RGB+D} videos,'' \emph{IEEE Transactions
  on Pattern Analysis and Machine Intelligence}, vol.~40, no.~5, pp.
  1045--1058, 2018.

\bibitem{tran_CVPR17}
L.~Tran, X.~Liu, J.~Zhou, and R.~Jin, ``Missing modalities imputation via
  cascaded residual autoencoder,'' in \emph{Proceedings of IEEE Computer Vision
  and Pattern Recognition}, 2017, pp. 1405--1414.

\bibitem{multiview_action2}
Z.~Cai, L.~Wang, X.~Peng, and Y.~Qiao, ``Multi-view super vector for action
  recognition,'' in \emph{Proceedings of IEEE Computer Vision and Pattern
  Recognition}, 2014, pp. 596--603.

\bibitem{kinect_v1v2}
D.~Pagliari and L.~Pinto, ``Calibration of kinect for {X}box {O}ne and
  comparison between the two generations of microsoft sensors,'' vol.~15, pp.
  27\,569--27\,589, 10 2015.

\bibitem{tran2018closer}
D.~Tran, H.~Wang, L.~Torresani, J.~Ray, Y.~LeCun, and M.~Paluri, ``A closer
  look at spatiotemporal convolutions for action recognition,'' in
  \emph{Proceedings of IEEE Computer Vision and Pattern Recognition}, 2018, pp.
  6450--6459.

\bibitem{dropout}
N.~Srivastava, G.~Hinton, A.~Krizhevsky, I.~Sutskever, and R.~Salakhutdinov,
  ``Dropout: a simple way to prevent neural networks from overfitting,''
  \emph{Journal of Machine Learning Research}, vol.~15, no.~1, pp. 1929--1958,
  2014.

\bibitem{ren2015faster}
S.~Ren, K.~He, R.~Girshick, and J.~Sun, ``Faster {R-CNN}: {T}owards real-time
  object detection with region proposal networks,'' \emph{arXiv preprint
  arXiv:1506.01497}, 2015.

\bibitem{swish}
P.~Ramachandran, B.~Zoph, and Q.~V. Le, ``Searching for activation functions,''
  \emph{arXiv preprint arXiv:1710.05941}, 2017.

\bibitem{carreira2018short}
J.~Carreira, E.~Noland, A.~Banki-Horvath, C.~Hillier, and A.~Zisserman, ``A
  short note about kinetics-600,'' \emph{arXiv preprint arXiv:1808.01340},
  2018.

\bibitem{zhang2016chinese}
J.~Zhang, W.~Zhou, C.~Xie, J.~Pu, and H.~Li, ``Chinese sign language
  recognition with adaptive {HMM},'' in \emph{Proceedings of IEEE International
  Conference on Multimedia and Expo}, 2016, pp. 1--6.

\bibitem{SLR_dataset}
O.~M. Sincan and H.~Y. Keles, ``{AUTSL}: {A} large scale multi-modal turkish
  sign language dataset and baseline methods,'' \emph{IEEE Access}, vol.~8, pp.
  181\,340--181\,355, 2020.

\bibitem{vaezijoze2019ms-asl}
H.~Vaezi~Joze and O.~Koller, ``Ms-asl: A large-scale data set and benchmark for
  understanding american sign language,'' in \emph{The British Machine Vision
  Conference}, 2019.

\bibitem{albanie2020bsl}
S.~Albanie, G.~Varol, L.~Momeni, T.~Afouras, J.~S. Chung, N.~Fox, and
  A.~Zisserman, ``Bsl-1k: Scaling up co-articulated sign language recognition
  using mouthing cues,'' in \emph{European Conference on Computer Vision},
  2020, pp. 35--53.

\bibitem{kinect2_acc}
C.~Amon, F.~Fuhrmann, and F.~Graf, ``Evaluation of the spatial resolution
  accuracy of the face tracking system for {K}inect for windows {V}1 and
  {V}2,'' in \emph{Proceedings of AAAI Conference on Artificial Intelligence},
  2014, pp. 16--17.

\bibitem{mmpose2020}
M.~Contributors, ``{OpenMMLab} pose estimation toolbox and benchmark,''
  \url{https://github.com/open-mmlab/mmpose}, 2020.

\bibitem{sun2019deep}
K.~Sun, B.~Xiao, D.~Liu, and J.~Wang, ``Deep high-resolution representation
  learning for human pose estimation,'' in \emph{Proceedings of IEEE Computer
  Vision and Pattern Recognition}, 2019, pp. 5693--5703.

\bibitem{denseflow}
S.~Wang, Z.~Li, Y.~Zhao, Y.~Xiong, L.~Wang, and D.~Lin, ``{denseflow},''
  \url{https://github.com/open-mmlab/denseflow}, 2020.

\bibitem{zach2007duality}
C.~Zach, T.~Pock, and H.~Bischof, ``A duality based approach for realtime
  {TV-L}1 optical flow,'' in \emph{Proceedings of Joint Pattern Recognition
  Symposium}.\hskip 1em plus 0.5em minus 0.4em\relax Springer, 2007, pp.
  214--223.

\bibitem{de2021isolated}
M.~De~Coster, M.~Van~Herreweghe, and J.~Dambre, ``Isolated sign recognition
  from rgb video using pose flow and self-attention,'' in \emph{Proceedings of
  the IEEE/CVF Conference on Computer Vision and Pattern Recognition}, 2021,
  pp. 3441--3450.

\bibitem{hu2020global}
H.~Hu, W.~gang Zhou, J.~Pu, and H.~Li, ``Global-local enhancement network for
  nmf-aware sign language recognition,'' \emph{ACM Transactions on Multimedia
  Computing, Communications, and Applications (TOMM)}, vol.~17, pp. 1 -- 19,
  2020.

\bibitem{hara2018can}
K.~Hara, H.~Kataoka, and Y.~Satoh, ``Can spatiotemporal {3D CNN}s retrace the
  history of {2D CNN}s and {ImageNet}?'' in \emph{Proceedings of IEEE Computer
  Vision and Pattern Recognition}, 2018, pp. 6546--6555.

\bibitem{kataoka2020would}
H.~Kataoka, T.~Wakamiya, K.~Hara, and Y.~Satoh, ``Would mega-scale datasets
  further enhance spatiotemporal {3D CNN}s?'' \emph{arXiv preprint
  arXiv:2004.04968}, 2020.

\end{thebibliography}
